\documentclass[sigconf]{acmart}
\renewcommand\footnotetextcopyrightpermission[1]{} 
\usepackage{booktabs} 

\setcopyright{none}

\begin{document}
\title{Convolutional neural networks pretrained on large face recognition datasets for emotion classification from video}

\author{Boris Knyazev}
\authornote{Currently at the University of Guelph.}
\orcid{1234-5678-9012}
\affiliation{%
	\institution{NTechLab}
	\city{Moscow} 
	\state{Russia} 
}
\email{b.knyazev@ntechlab.com}

\author{Roman Shvetsov}
\orcid{1234-5678-9012}
\affiliation{%
	\institution{NTechLab}
	\city{Moscow} 
	\state{Russia} 
}
\email{r.shvetsov@ntechlab.com}

\author{Natalia Efremova}
\affiliation{%
	\institution{University of Oxford}
	\streetaddress{}
	\city{Oxford} 
	\state{UK} 
}
\email{natalia.efremova@sbs.ox.ac.uk}

\author{Artem Kuharenko}
\affiliation{%
	\institution{NTechLab}
	\city{Moscow} 
	\state{Russia} 
}
\email{a.kuharenko@ntechlab.com}

\renewcommand{\shortauthors}{B. Knyazev et al.}

\begin{abstract}
In this paper we describe a solution to our entry for the emotion recognition challenge EmotiW 2017. We propose an ensemble of several models, which capture spatial and audio features from videos. Spatial features are captured by convolutional neural networks, pretrained on large face recognition datasets.
We show that usage of strong industry-level face recognition networks increases the accuracy of emotion recognition.
Using our ensemble we improve on the previous best result on the test set by about 1~\%, achieving a 60.03~\% classification accuracy without any use of visual temporal information.
\end{abstract}

%
%



\maketitle

\section{Introduction}

Emotion recognition potentially has many applications in academia and industry, and emotional intelligence is an important part of artificial intelligence. However, in contrast to such tasks as face recognition (FR), emotion recognition has not yet become so widespread. We believe that the reason for this is the fact that emotion recognition is much harder and requires more research and efforts to gain success. Face recognition is also hard, but training data with clean ground truth labels can be collected easier and benchmarks are usually objective (i.e. we know the identity).
In emotion recognition, there is a lack of understanding and the agreement of what the labels should be. This can be proved by recent appearance of datasets with compound emotions~\cite{benitez2017emotionet} or with dominant and 
complementary emotions \cite{lusi2017joint}. There is also a lack of training data due to difficulty of collecting rare emotions (how often do you clearly show fear?).

Emotion recognition from video is also more difficult than general video recognition. State of the art methods such as (Improved) Dense Trajectories~\cite{wang2013action} or 3D convolutional neural networks, which typically show a 70-90 \% accuracy on video datasets, fail to provide results above 40 \% on emotion datasets~\cite{yao2016holonet,fan2016video}.

The emotion recognition challenges, in particular EmotiW 2017 and its predecessors~\cite{dhall2012collecting, dhall2016emotiw}, allow to boost the progress in this area by providing data and benchmarks for training and evaluating novel methods. Once emotions can be recognized reliably and well understood, it can provide the same or even more benefits than face recognition. Due to the presence of concealed emotions it can lead to even more benefits because humans need expert and rare knowledge to recognize concealed emotions, while machines could potentially perform this task easily opening up new research areas as well as new privacy challenges analogously to face recognition.

In this work, we attempt to further contribute to the field of emotion recognition by presenting our solution to the fifth Emotion Recognition in the Wild Challenge (EmotiW) 2017, in particular to its  audio-video emotion recognition sub-challenge.

Recent NIST reports show that our FR networks perform as state-of-the-art or better in most benchmarks~\cite{nist2017grother} and, in this work, we employ them for emotion recognition by fine-tuning them on the emotion datasets.
Similarly to other video recognition challenges~\cite{abu2016youtube}, we make features extracted from facial images of all frames publicly available.

\section{Experiments}

\subsection{Networks}

We experiment with four convolutional neural networks: VGG-Face~\cite{parkhi2015deep} and three proprietary state of the art face recognition networks which we notate as FR-Net-A, FR-Net-B and FR-Net-C. Compared to VGG-Face, which is trained on 2600 individuals with around 3 million images, our networks are trained on a much larger data volume (Table ~\ref{tab:data}), which make them more powerful both for face and emotion recognition tasks.


\subsection{Datasets}

In this work we use two emotion datasets to train our models (Table~\ref{tab:data}): FER2013~\cite{goodfellow2013challenges} and data of this challenge EmotiW 2017. The FER2013 dataset was also used in the previous year's winning method~\cite{fan2016video}. It consists of 28709 training, 3589 validation and 3589 test images. We fine-tune our models using only a training set.

In the audio-video emotion recognition sub-challenge of Emotiw 2017 there are the same training (773) and validation (383) videos as in the EmotiW 2016 version, but this year 60 new test videos were added with 653 in total.  

\subsection{Pipeline}

\subsubsection{Face detection}
To extract and align faces both from images of FER2013 and EmotiW video frames we use the dlib face detector~\cite{dlib09}. If a face was not found on the frame, the entire frame is passed to the network. To limit such cases, we apply a low face detection threshold.

\subsubsection{Frames feature extraction}
In all experiments we follow the same pipeline to obtain results on EmotiW validation videos.
First, features for all frames are computed using all four networks. For VGG-Face, we empirically choose fc6 features (4096 dimensional), while for other networks we use 1024 dimensional features of the layer before the last one.

\subsubsection{Frame-level feature aggregation}

Motivated by results of the previous year~\cite{bargal2016emotion}, we compute features of videos with one or several aggregation functions (e.g., mean, standard deviation) (Table~\ref{tab:features}) followed by rescaling to range $[-1,1]$ and concatenation (e.g., in case of VGG-Face if we compute mean and std features, then features are 8192 dimensional). In this work, rootsift normalization ($\mathrm{sign}(\mathbf{x})\sqrt{| \mathbf{x} |/ \| \mathbf{x} \|_1 }$)~\cite{arandjelovic2012three} and global standardization is also applied to concatenated features.

\subsubsection{Classification}

A linear SVM was trained on training data (one SVM per network) in case of reporting validation accuracy and, as in~\cite{fan2016video}, on training plus validation data in case of test accuracy. The regularization constant of SVMs is found by 5-fold cross-validation.

\subsubsection{Ensembling with audio features}
We compute 1582 dimensional audio features using the Opensmile library~\cite{eyben2010opensmile}. We train a linear SVM in this case as well, so that our ensemble averages scores of 5 SVMs in total.

\subsection{FER2013 fine-tuning}

All four networks are fine-tuned on FER2013 by replacing the last 1-2 fully connected layers with new ones (for VGG we only replace the classification layer) and then training all layers for 30k iterations with Nesterov SGD with the following parameters: learning rate 0.0001-0.0005, weight decay 0.0005, momentum 0.9, batch size 32 and polynomial learning rate decay. 
The networks achieved 70-72~\% accuracy on FER2013 validation data, but we did not pay much attention to these results, because the relationship between the FER2013 and validation accuracies was not always straightforward.

\begin{table}
	\caption{Training datasets used in this work. For our face recognition data we only provide the number of images.}
	\label{tab:data}
	\begin{tabular}{ccl}
		\toprule
		Dataset & \#classes & \#images \\
		\midrule
		FER 2013 & 7 emotions & 35k \\
		EmotiW 2017 & 7 emotions & 50k frames \\
		\midrule
		VGG-Face data & 2600 persons & 3m \\
		Our face recognition data & - & 50m \\
		\bottomrule
	\end{tabular}
\end{table}

%
%
%


As expected, fine-tuning (FT) on FER2013 boosts performance on EmotiW in all cases (Table~\ref{tab:models}). Before FT better FR models usually have worse performance in the emotion recognition task, because face recognition should be emotion invariant, but after FT the results invert. For instance, FR-Net-A and FR-Net-B have the same architecture, but the latter was pretrained on much larger face recognition data. As a result, FR-Net-B is the worst before fine-tuning, but the best afterwards. This experiment confirms that pretraining on larger FR data positively affects emotion recognition accuracy.

\begin{table}
	\caption{Model comparison using frames feature averaging on the validation set before and after fine-tuning (FT) on FER2013. Best results in each column are \textbf{in bold}.}
	\label{tab:models}
	\begin{tabular}{lccc}
		\toprule
		Model & Before FT & After FT & \# features\\
		\midrule
		VGG-Face & \textbf{37.9} & 48.3 & 4096\\
		FR-Net-A & 33.7 & 44.6 & 1024 \\ 
        FR-Net-B & 33.4 & \textbf{48.8} & 1024 \\ 
        FR-Net-C & 37.6 & 45.2 & 1024 \\ 
        \midrule
        Audio features & 35.0 & - & 1582 \\
		\bottomrule
	\end{tabular}
\end{table}













\subsection{Feature comparison}
To improve STAT features used in~\cite{bargal2016emotion}, which include mean, standard deviation, minimum and maximum features we first performed an ablation study and removed the \textit{max} features (Table~\ref{tab:features}). This appeared to be important for improving generalization on the test set. 

Afterwards, we found that frame-based augmentation in the form of horizontal flipping, rotation and scaling usually improves features except for VGG-Face. We compute 18 transformations per frame and average features of these transformations.

We then tried to add spectral features by computing the 1 dimensional Fourier transform (fft) for each neuron and then taking the average of that. For instance, for VGG-Face for one video we have 4096 dimensional fft features. These features significantly improved performance on the validation set (Table~\ref{tab:features}), however, due to the limit of test submissions, we were not able to evaluate this and several other features. 

\begin{table}
	\caption{Feature comparison on the validation set for models fine-tuned on FER2013. STAT* is STAT without \textit{max}. ** - our best submission. A - augmentation. Ensemble: VGG-Face + FR-Net-A+B+C + Audio. Best results in each row are \textbf{in bold}.}
	\label{tab:features}
    \setlength{\tabcolsep}{1pt}
	\begin{tabular}{lcccccc}
		\toprule
		Model & Mean & STAT & STAT* & STAT*+A & STAT*+fft & STAT*+fft+A \\
		\midrule
		VGG-Face & 48.3 & 52.2 & 52.5 & 50.4 & \textbf{53.0} & 50.7 \\
		FR-Net-A & 44.6 & 47.8 & 47.5 & 49.3 & 48.3 & \textbf{49.6} \\ 
        FR-Net-B & 48.8 & 52.5 & 52.2 & 52.5 & 53.3 & \textbf{53.5} \\ 
        FR-Net-C & 45.2 & 45.2 & 45.7 & \textbf{53.0} & 45.2 & 52.5 \\ 
        \midrule
        Ensemble & 52.7 & 55.1 & 54.8 & 56.4** & 56.4 & \textbf{56.7} \\
		\bottomrule
	\end{tabular}
\end{table}

\subsection{Class distribution}
In this challenge we noticed that fitting the validation set was not useful or was even harmful, because it differs quite a lot from the test set.
Specifically, the validation set is relatively balanced, i.e. each emotion has about the same number of samples, while according to the confusion matrix (provided for submissions) of the test set it is imbalanced with many more samples of happy, neutral and angry emotions (Table~\ref{tab:classes}). Therefore, if during fitting of the validation set the model that predicts all emotions equally well will be chosen, that would decrease performance on the test set, because on the test set not all emotions are equally important.

This observation enables us to significantly improve results by weighting scores of emotions according to the square root of the observed test set frequencies. As a result, our model minimizes the presence of disgust and surprise emotions in order to  improve recognition of happy and neutral facial expressions and to achieve better classification accuracy of 60.03\% overall (Table~\ref{tab:ensembles}).

\begin{table}
  \caption{Comparison of our ensembles with other results. Ensemble: VGG-Face + FR-Net-A+B+C + Audio.}
  \label{tab:ensembles}
  \begin{tabular}{lcc}
    \toprule
	Model & Val acc & Test acc \\
    \midrule
    Baseline~\cite{dhall2016emotiw} & 38.81 & 40.47 \\
    Previous year best results & 59.42~\cite{bargal2016emotion} & 59.02~\cite{fan2016video} \\
    \midrule
    Ensemble & 54.83 & - \\
    Ensemble + A & 56.40 & 54.98 \\
    Ensemble + A + class weights & 48.30 & 60.03 \\
  \bottomrule
\end{tabular}
\end{table}

\begin{figure}
\label{fig:confusion}
\includegraphics[scale=0.4]{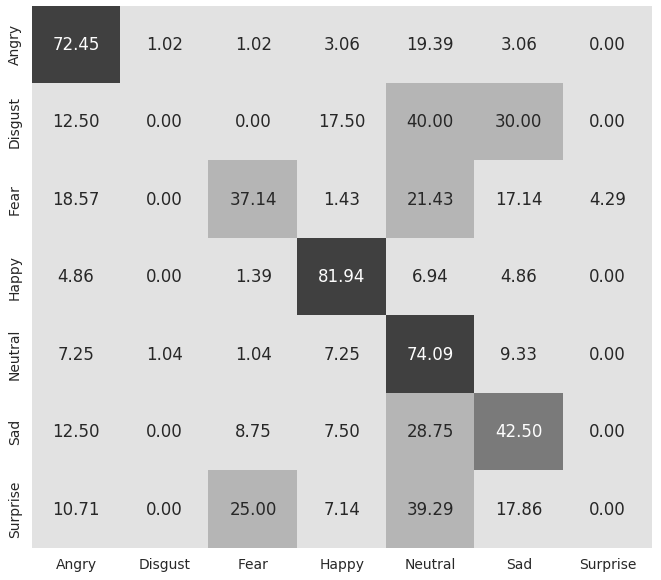}
\caption{The confusion matrix of our final predictions on the test set.}
\end{figure}

\begin{table}
  \caption{Training, validation and test sets class distribution in EmotiW 2017.}
  \label{tab:classes}
  \setlength{\tabcolsep}{4pt}
  \begin{tabular}{lcccccccc}
    \toprule
	 & An & Di & Fe & Ha & Ne & Sa & Su & Total \\
    \midrule
    Train & 133 & 74 & 81 & 150 & 144 & 117 & 74 & 773 \\
    Validation & 64 & 40 & 46 & 63 & 63 & 61 & 46 & 383  \\
    Test & 98 & 40 & 70 & 144 & 193 & 80 & 28 & 653  \\
    \midrule
    Class weights & 0.15 & 0.10 & 0.13 & 0.19 & 0.21 & 0.14 & 0.08 & 1 \\
  \bottomrule
\end{tabular}
\end{table}

\section{Frame shuffle augmentation for LSTM}
In this experiment, we trained a Long Short-Term Memory (LSTM) by randomly shuffling the order of frames during training (Table~\ref{tab:lstms}), which can be seen as a form of augmentation. Although it sounds counterintuitive, we achieve considerable accuracy gain compared to training on frames in the original order (during inference the order is fixed). Our results imply that each video in this task is not a \textit{sequence} of video frames, but rather a \textit{set} of frames. 
Manual examination of videos was consistent with our finding, because after shuffling frames of some video we still could (or still couldn't) recognize the correct emotion.
LSTMs are trained on top of our best FR network analogously to~\cite{fan2016video} on full sequences.
However, we were not able to improve our performance on the test set by adding LSTMs. 

\begin{table}
	\caption{For this task the order of frames is not very important, so the videos can be seen as unordered sets of frames.}
	\label{tab:lstms}
	\begin{tabular}{lc}
		\toprule
		Model & Val acc \\
		\midrule
		LSTM + FR-NET-B & 46.48 \\
		LSTM + FR-NET-B + frame shuffling & 50.39 \\
		\bottomrule
	\end{tabular}
\end{table}

\section{Conclusions}

In this work, we present an ensemble of models that achieves better emotion classification accuracy than the previous year's winner. We rely on strong face recognition convolutional networks which can be easily fine-tuned to perform the emotion recognition task. Audio features are also used to complement our models with an additional modality. We make frame level features computed with our networks publicly available to help the research community by reducing the task of emotion recognition from video to learning from high level features. 

\begin{acks}
The authors would like to thank other members of the NTechLab team for useful advice and technical support.
\end{acks}

\bibliographystyle{ACM-Reference-Format}
\bibliography{sigproc} 

\end{document}